\documentclass[letterpaper, 10 pt, conference]{ieeeconf}  

\IEEEoverridecommandlockouts                              

\usepackage{graphicx} 
\usepackage{color}
\usepackage{booktabs}
\usepackage{multirow} 
\usepackage{caption}
\usepackage{lipsum}
\usepackage{tabularx}
\usepackage{float}
\usepackage{stfloats}
\usepackage{makecell}
\usepackage{amsmath,amsfonts}
\usepackage{algorithmic}
\usepackage{url}
\usepackage{cite}
\usepackage[colorlinks=true, linkcolor=blue, citecolor=blue, urlcolor=blue]{hyperref}
\usepackage{adjustbox}
\usepackage[table]{xcolor}

\usepackage{balance}

\title{\LARGE \bf
LLM-RG: Referential Grounding in Outdoor Scenarios using Large Language Models
}

\author{Pranav Saxena$^{1,2}$ Avigyan Bhattacharya$^{2}$ Ji Zhang$^{2}$ Wenshan Wang$^{2}$
\thanks{$^{1}$Pranav Saxena is with Birla Institute of
Technology and Science Pilani, K.K Birla Goa Campus, Goa, India
        {\tt\small f20220257@goa.bits-pilani.ac.in}}%
\thanks{$^{2}$The authors are with Carnegie Mellon University, Robotics
Institute, Pittsburgh, PA.
        {\tt\small \{psaxena2,avigyanb, zhangji,
wenshanw\}@andrew.cmu.edu}}%
}

\begin{document}

\maketitle
\thispagestyle{empty}
\pagestyle{empty}

\begin{abstract}

Referential grounding in outdoor driving scenes is challenging due to large scene variability, many visually similar objects, and dynamic elements that complicate resolving natural-language references (e.g., “the black car on the right”). We propose LLM-RG, a hybrid pipeline that combines off-the-shelf vision–language models for fine-grained attribute extraction with large language models for symbolic reasoning. LLM-RG processes an image and a free-form referring expression by using an LLM to extract relevant object types and attributes, detecting candidate regions, generating rich visual descriptors with a VLM, and then combining these descriptors with spatial metadata into natural-language prompts that are input to an LLM for chain-of-thought reasoning to identify the referent’s bounding box. Evaluated on the Talk2Car benchmark, LLM-RG yields substantial gains over both LLM and VLM-based baselines. Additionally, our ablations show that adding 3D spatial cues further improves grounding. Our results demonstrate the complementary strengths of VLMs and LLMs, applied in a zero-shot manner, for robust outdoor referential grounding.

\end{abstract}

\section{INTRODUCTION}

Enabling autonomous systems to ground referring expressions to real-world entities in complex settings is a crucial step toward safe and natural interactions with humans. In contrast to indoor settings, which have been the focus of most prior works, outdoor scenes pose distinct challenges due to larger scales, greater object diversity, and more complex, dynamic environments like roads and intersections. Referential expressions in this setting frequently rely on high-level attributes (such as \textit{color}, \textit{orientation}, or \textit{type}) and relative spatial relations (such as ``on the right'' or ``behind the van''), which are harder to resolve than the structured references typically found indoors.  

In recent years, significant progress has been made in grounding referential language within indoor environments. Large-scale 3D datasets such as Matterport3D~\cite{Matterport3D}, ScanNet~\cite{dai2017scannet}, and HM3D~\cite{ramakrishnan2021hm3d} have enabled tasks including visual grounding, embodied instruction following, and object-goal navigation. Methods developed on these datasets often combine object detection with geometric or spatial reasoning modules~\cite{yu2018mattnet,chen2020scanrefer,liu2019improving}, or leverage pretrained language models to link natural language queries to structured scene graphs~\cite{shridhar2020alfred,zemskova20253dgraphllmcombiningsemanticgraphs}. More recently, large multimodal models have been explored for zero-shot grounding in 3D indoor spaces~\cite{fang2024transcrib3d3dreferringexpression,zantout2025sort3dspatialobjectcentricreasoning}. These approaches have shown strong performance in resolving references to small household objects and reasoning about relations such as ``next to the chair'' or ``on top of the table.'' However, they remain heavily tuned to the relatively constrained and repetitive structure of indoor scenes, where object categories are limited, contexts are predictable, and the variability of natural language references is narrower.  

\begin{figure}[t]
    \centering
    \includegraphics[width=0.5\textwidth]{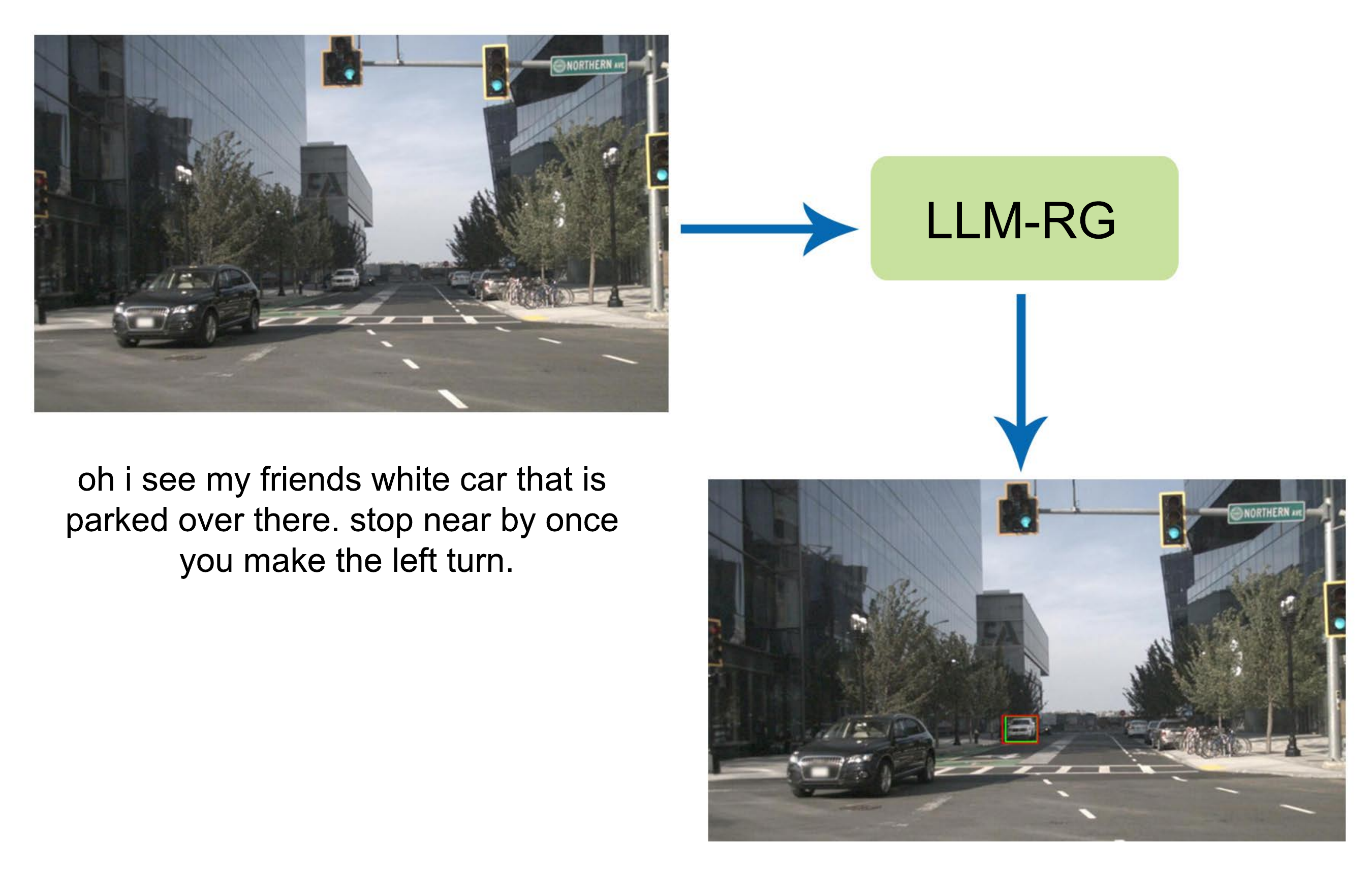}
    \caption{An example output of LLM-RG on a scene from Talk2Car. Red box denotes Ground Truth bounding box from Talk2Car, green box denotes the predicted bounding box using LLM-RG.}
    \vspace{-1em}
    \label{fig:arch}
\end{figure}

In contrast, outdoor referential grounding has received considerably less attention, even though it is essential for applications in autonomous driving, mobile robotics, and delivery systems. Outdoor environments are inherently more complex: they contain a larger and more open-ended vocabulary of objects (e.g., cars, trucks, pedestrians, bicycles, traffic lights), involve greater scene variability (e.g., urban streets, intersections, crosswalks, parking lots), and are subject to dynamic changes such as moving vehicles or occlusions. Furthermore, outdoor language queries tend to be more diverse and ambiguous, often requiring fine-grained disambiguation across multiple similar objects (e.g., ``the black car on the right'' when several black cars are present) or reasoning over higher-level semantics (e.g., ``the car waiting at the stop sign''). Datasets such as Talk2Car~\cite{deruyttere2019talk2car} address this gap by providing natural language commands linked to visual driving scenes, but methods specifically designed for outdoor referential grounding remain scarce. 

To address these challenges, we propose to leverage recent advances in vision-language models (VLMs) for extracting fine-grained object attributes and large language models (LLMs) for reasoning over natural language queries. We evaluate this approach on the Talk2Car dataset, which provides a realistic and challenging benchmark for outdoor referential grounding in driving scenarios. 

Our work, based on SORT3D~\cite{zantout2025sort3dspatialobjectcentricreasoning}, introduces these three key contributions -
\begin{enumerate}
    \item We present a novel pipeline that combines VLM-based object attribute extraction with LLM-based reasoning for outdoor referential grounding.
    \item We show that our approach works without any task-specific fine-tuning, making it deployable across unseen datasets and robotic setups.
    \item We provide an extensive evaluation demonstrating the effectiveness of this hybrid approach and highlight its potential for natural human-vehicle interaction in real-world contexts.

\end{enumerate}

\begin{figure*}[t]
    \centering
    \includegraphics[width=0.85\textwidth]{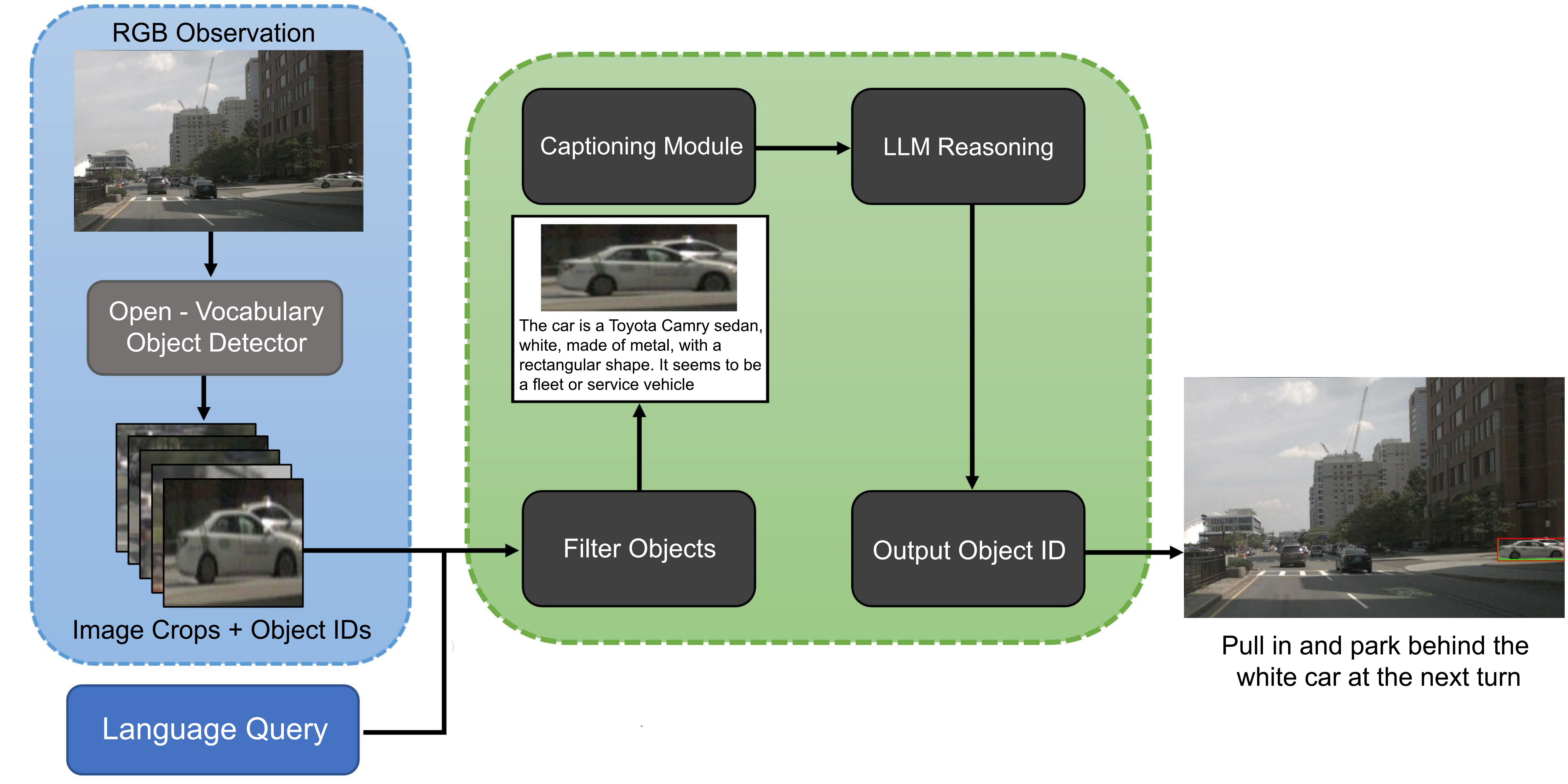}
    \caption{\textbf{Architecture of LLM-RG:} 
    \textbf{(A)} A large language model (LLM) processes the referring expression to identify relevant object types and attributes, generating a shortlist of candidate objects.
    \textbf{(B)} MMDetection is used to detect objects and obtain 2D bounding boxes.
    \textbf{(C)} Object crops for each detection are extracted and passed to a vision-language model (VLM) which provides fine-grained descriptions of each candidate object, capturing properties such as color, type, orientation, and contextual details.
    \textbf{(D)} The LLM combines object IDs, spatial locations, and object descriptions to reason over the referring expression and identify the bounding box of the target object.
    }
    \vspace{-1em}
    \label{fig:arch}
\end{figure*}

\section{Related Work}

\subsection{Referential Grounding in Indoor Scenes}

Referential grounding in indoor environments has become a major focus, aiming to connect natural language with complex 3D scenes. \textbf{Scene-LLM} \cite{fu2024scenellmextendinglanguagemodel} integrates large language models with 3D visual data for tasks like scene captioning, object identification, and interactive planning. \textbf{IRef-VLA} \cite{zhang2025irefvlabenchmarkinteractivereferential} introduces a benchmark with over 11.5K scanned rooms and 4.7M referential statements for precise language grounding to objects and actions. \textbf{3D-GRAND} \cite{yang2024_3D_GRAND} provides a large-scale 3D-text dataset with 40K+ scenes and 6.2M instructions, supporting object reference, spatial reasoning, and training of 3D-LLMs.

While indoor benchmarks have advanced referential grounding, our focus is on outdoor driving scenes, which present greater diversity, dynamics, and ambiguity relevant to our application.

\subsection{Referential Grounding in Outdoor Scenes}

Although relatively underexplored, referential grounding in outdoor environments introduces distinct challenges stemming from the large scale and variability of scenes. \textbf{LidaRefer} \cite{baek2025lidarefercontextawareoutdoor3d} addresses these challenges with a transformer-based 3D visual grounding framework tailored for large-scale outdoor scenes, tackling issues such as high computational demands and ambiguous object identification in sparse LiDAR point clouds. Meanwhile, \textbf{Grounded 3D-LLM} \cite{chen2024grounded3dllmreferenttokens} introduces the use of scene referent tokens to reference 3D scenes, enabling the processing of interleaved 3D and textual data sequences and unifying multiple 3D vision tasks within a generative framework.
While these methods rely on extensive training to adapt to outdoor domains, our pipeline remains training-free, enabling flexible zero-shot deployment across new datasets and scenarios.

\subsection{LLMs and VLMs for Referential Grounding}

The integration of large language models (LLMs) and vision-language models (VLMs) has led to significant progress in referential grounding across both indoor and outdoor environments. Recent approaches leverage the complementary strengths of language understanding and visual perception to address increasingly complex grounding tasks. \textbf{SpatialVLM}~\cite{Chen_2024_CVPR} introduces a vision-language model capable of interpreting spatial relationships directly from language, without relying on explicit 3D scene representations, enabling the resolution of spatial queries through purely linguistic and visual cues. \textbf{GLaMM}~\cite{rasheed2024glammpixelgroundinglarge} demonstrates robust multimodal capabilities by supporting scene-level understanding, region-level interpretation, and pixel-level grounding, effectively addressing a broad range of visual grounding challenges across different granularities. Extending these ideas into the 3D domain, \textbf{ScanReason}~\cite{zhu2024scanreasonempowering3dvisual} proposes a 3D visual grounding framework that combines multimodal LLMs with a visual-centric reasoning module and a 3D grounding module, enhancing geometric understanding and fine-grained object localization within complex 3D scenes.

Unlike prior work, our framework adopts a modular chaining strategy, where a VLM generates fine-grained visual descriptions and an LLM performs symbolic reasoning, unifying perception and reasoning in a zero-shot manner.

\section{Methodology}


In this work, we present \textbf{LLM-RG}, a hybrid LLM-VLM pipeline for outdoor referential grounding, specifically designed to handle the scale, diversity, and dynamic nature of real-world driving scenes. The system takes a driving scene and a free-form referring expression as input and outputs the bounding box corresponding to the referenced object. By leveraging large language models for reasoning and vision-language models for fine-grained attribute extraction, our approach enables robust zero-shot grounding in challenging urban environments.

Figure \ref{fig:arch} provides an overview of the proposed pipeline. Each component of the system is described in detail in the following subsections.

\subsection{Object Detection and Localization}

Given an RGB image paired with a referential statement, we first use a Large Language Model (LLM) to extract the relevant object categories mentioned in the query. This acts as a textual filter, narrowing down candidate objects in complex scenes. For instance, considering the referential statement “Park near the car under the tree,” the extracted relevant objects are [``car", ``tree"].

The filtered categories are then passed to an open-vocabulary object detector \cite{liu2023grounding}, which predicts bounding boxes and class labels for each candidate object. This combination of textual filtering and visual detection allows the system to focus on relevant objects efficiently.

\subsection{Object Feature Extraction}

\begin{figure}[!htp]
   \includegraphics[width=\linewidth]{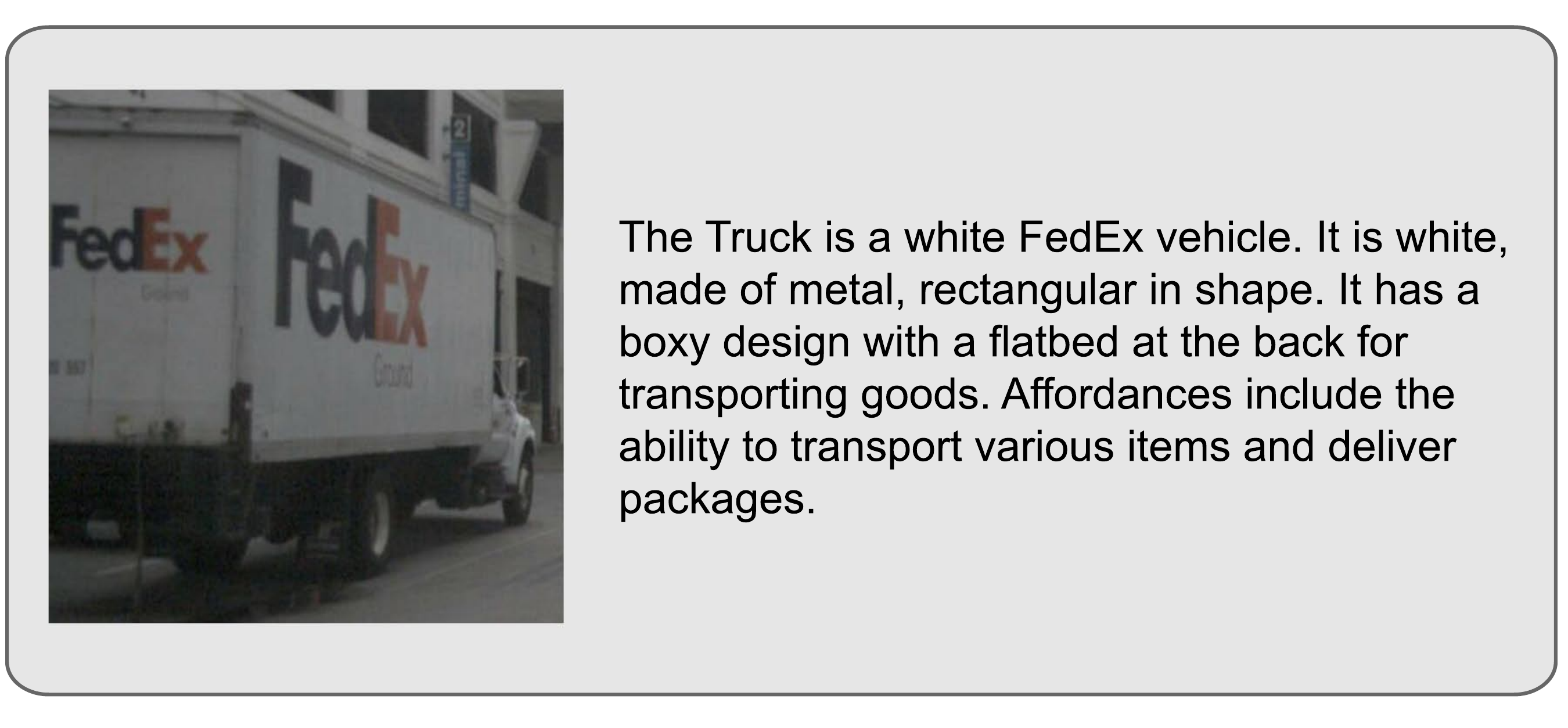}
    \caption{Example of an object caption from a VLM that includes fine-grained attributes to be used for further reasoning.}
    \label{fig:caption}
\end{figure}

For each detected object, we crop the corresponding image region and associate it with its class label. We then use a large vision-language model ~\cite{2409.12191} to generate fine-grained attributes such as color, material, shape, affordances, and other meaningful attributes. These object descriptors provide richer semantic information than bounding boxes alone, enabling the system to capture details critical for disambiguating similar objects. This step mirrors human perception, where attributes help distinguish between multiple candidate objects in a scene. An example caption is shown in Figure \ref{fig:caption}.

\subsection{Prompt Construction and LLM Reasoning}

The extracted object attributes, bounding box coordinates, and class labels are formatted into a natural language prompt for the LLM \cite{mistral2024} to reason about the scene, represented as \textit{[id, 'name', 'caption', [x,y]]}. To improve reasoning, we include example input-output pairs, which we find helps increase the reasoning significantly, and instruct the LLM to use chain-of-thought reasoning. This structured prompt enables the LLM to interpret visual and spatial information in textual form, thereby providing the necessary context to accurately identify the correct referent from the candidate objects.

The LLM processes the prompt to identify the object that best matches the referential statement. Once the LLM outputs the object ID, it is mapped back to the corresponding bounding box. 

\section{Experiments and Results}

\subsection{Evaluation}

\textbf{Dataset: }We evaluate on the Talk2Car dataset, which provides real-world driving scenes from nuScenes\cite{caesar2020nuscenesmultimodaldatasetautonomous} paired with free-form referring expressions and corresponding ground-truth bounding boxes.

\textbf{Metrics: } We report Accuracy (Acc@0.5), the percentage of predictions with Intersection over Union (IoU) $\geq$ 0.5 with the ground truth.

\begin{figure*}[t]
    \centering
    \includegraphics[width=1.0\textwidth]{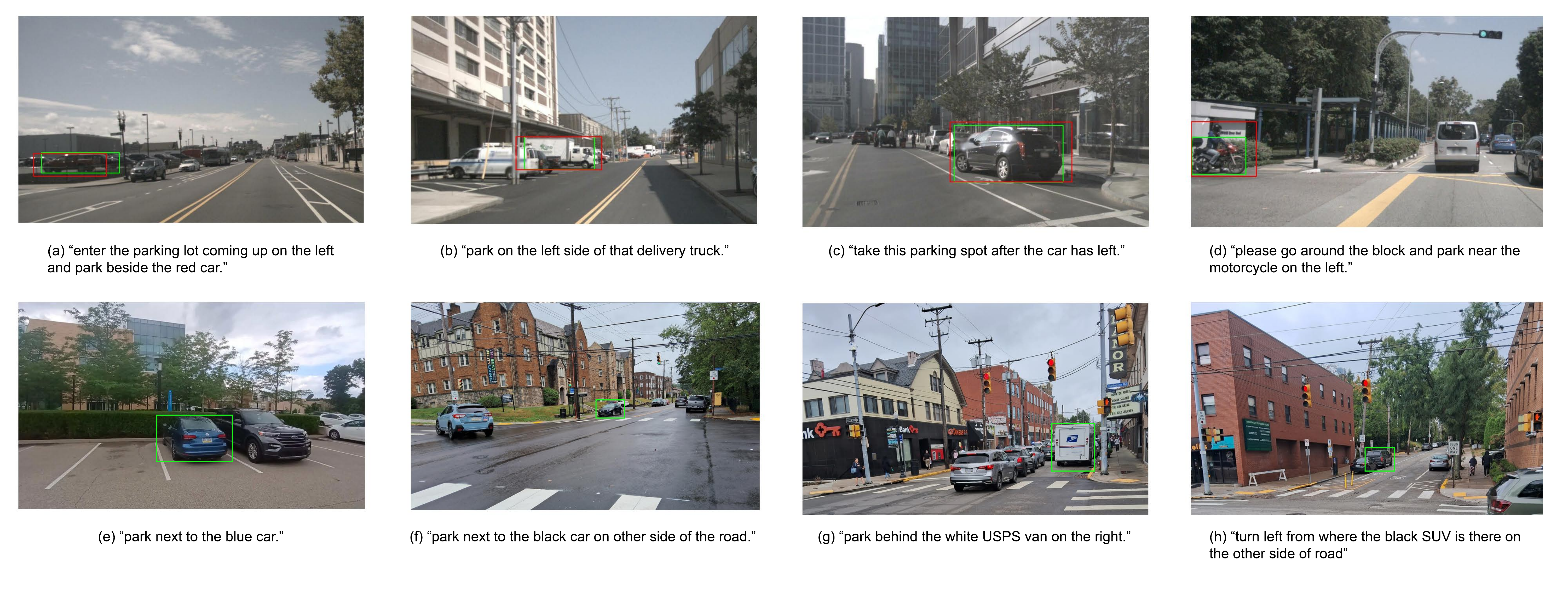}
    \caption{Qualitative results of LLM-RG on Talk2Car (first row) and mecanum robot (second row). Red box denotes Ground Truth bounding box from Talk2Car, green box denotes the predicted bounding box using LLM-RG.}
    \vspace{-1em}
    \label{fig:t2c}
\end{figure*}

\subsection{Baselines}

We compare our results against a recent work and against different baselines that we design:

(i) \textbf{LLM-Wrapper~\cite{cardiel2025llmwrapperblackboxsemanticawareadaptation}: }It adapts off-the-shelf VLMs for referring expression comprehension in a black-box setting. The method converts candidate bounding boxes from a VLM into textual prompts and uses an LLM to reason over them, selecting the most relevant box. We compare against its zero-shot version.

(ii) \textbf{Naive-VLM: }We utilize a VLM (Gemini 2.5-flash) by passing the RGB image and the referential statement directly. The model outputs a bounding box without any additional reasoning or context. This serves as a simple baseline for zero-shot comprehension.

(iii) \textbf{Image Crops + VLM: }We crop the image into object regions and assign each region an object ID. These crops with IDs are passed to the VLM, which is asked to output the best matching object ID. We then compare the IoU of the predicted bounding box corresponding to the best object ID with the ground-truth bounding box.

(iv) \textbf{Image with bounding box + captions + VLM: }We pass the entire image with detected bounding boxes along with object IDs and captions to a VLM (Gemini 2.5-flash) and ask it output the best object ID. We then compare the IoU of the predicted bounding box corresponding to the best object ID with the ground-truth bounding box.

\begin{table}[H]
    \centering
    \renewcommand{\arraystretch}{1.2}
    \resizebox{\linewidth}{!}{%
    \begin{tabular}{l c c}
        \toprule
        Method & val & test \\
        \midrule
        Naive-VLM                   & 37.13 & 39.23 \\
        Image Crops + VLM            & 47.32 & 48.11 \\
        Image with bounding box + Captions + VLM & 63.12 & 63.78 \\
        LLM-Wrapper (Zero-Shot)                  & 55.37  & 58.44 \\
        LLM-RG                       & \textbf{64.72} & \textbf{67.91} \\
        \bottomrule
    \end{tabular}%
    }
    \caption{Results of LLM-RG on Talk2Car Dataset.}
    \label{tab:results}
\end{table}

\subsection{Ablation Study}

We perform an ablation study \ref{tab:ablation} to evaluate the impact of incorporating 3D spatial information on reasoning performance. Specifically, we consider the following variants:

(i) We extend the Talk2Car framework to the NuScenes dataset by projecting 2D bounding boxes onto the corresponding LiDAR point cloud to obtain partial 3D bounding boxes. The centroid of each resulting 3D box is then used to estimate the approximate 3D location of the object.

(ii) We directly utilize the ground-truth 3D bounding boxes provided by NuScenes, bypassing the detection and projection steps, and perform reasoning based on these boxes.

We observe that having 3D Spatial Information helps improve reasoning significantly

\subsection{Qualitative Results}

We show the qualitative results of LLM-RG on Talk2Car dataset, as well as on a real-life mecanum robot, highlighting its zero-shot adaptability on a new robotic setup (Fig. \ref{fig:t2c}).

\vspace{0.6em}
\begin{table}[H]
    \centering
    \renewcommand{\arraystretch}{1.2}
    \resizebox{\linewidth}{!}{%
    \begin{tabular}{l *{2}{c}}
        \toprule
        Method  & Accuracy \\
        \midrule
        LLM-RG & 64.72 \\
        LLM-RG + LIDAR          & 70.23 \\
        LLM-RG + GT Bounding boxes from NuScenes        & 77.93\\
        \bottomrule
    \end{tabular}
    }
    \caption{Comparison of reasoning across different methods}.
    \vspace{-1em}
    \label{tab:ablation}
\end{table}

\section{Conclusion And Future Work}

We present LLM-RG, a hybrid referential grounding pipeline for outdoor driving scenes that leverages off-the-shelf vision-language models for detailed visual description and large language models for flexible, symbolic reasoning. Operating zero-shot without task-specific fine-tuning, LLM-RG parses free-form referring expressions into object-level attributes, generates candidate detections, and converts visual descriptors and spatial metadata into natural language prompts, framing grounding as a language-guided selection problem. Evaluations on the Talk2Car benchmark show this modular approach substantially improves over simple VLM baselines and related methods.

Our results highlight two key strengths: the complementary capabilities of VLMs, for extracting fine-grained image attributes, and LLMs, for compositional reasoning over them; and the practicality of a modular pipeline that can exploit black-box models without end-to-end retraining. These features make LLM-RG well-suited for outdoor driving scenarios requiring fast domain adaptation and interpretable outputs. Our ablations further reveal that incorporating 3D spatial information significantly improves grounding accuracy.

In future work, we plan to incorporate richer multimodal signals, such as depth maps and radar, to improve robustness in real-world driving. We also aim to extend LLM-RG to dynamic environments by integrating object tracking and temporal reasoning, enabling reliable operation in scenes with moving objects and complex interactions.

\bibliographystyle{unsrt}
\nocite{*}
\bibliography{references}

\end{document}